# TRACE: A Multi-Layer Benchmark for Human–AI–Controller Coordination Under Drift and Failure

Joshua Zuniga
*College of Computing and Software Engineering*
*Kennesaw State University*
*Marietta, GA, USA*
*jzuniga3@students.kennesaw.edu*

Srinivasan Subramanian
*College of Computing and Software Engineering*
*Kennesaw State University*
*Marietta, GA, USA*
*ssubram7@students.kennesaw.edu*

Ramya Madhuri Narapureddy
*College of Computing and Software Engineering*
*Kennesaw State University*
*Marietta, GA, USA*
*rnarapur@students.kennesaw.edu*

Md Abdullah Al Hafiz Khan
*College of Computing and Software Engineering*
*Kennesaw State University*
*Marietta, GA, USA*
*mkhan74@kennesaw.edu*

***Abstract*—Modern cyber-physical and AI-assisted systems couple human operators, AI decision modules, and automated controllers in a single control loop, so trustworthiness depends on the whole loop, not any one model. Yet no standard benchmark captures time-aligned, multi-layer traces of how drift and failures propagate across these layers, so we cannot diagnose where coordination breaks down, why, or how to recover. This paper targets one facet of that gap: drift, a deviation that can originate in any stack layer and that conventional single-modality monitoring cannot localize to a layer or pin to an onset time. We construct a benchmark by injecting controlled drift into traces derived from ALFRED, a grounded-instruction benchmark for everyday household tasks, yielding 1,918 drifted traces. Each trace is a time-aligned sequence of per-step records across five execution layers (state, observation, decision, rules, control), labeled with the drift type, affected layer, onset time, responsible actor, and causal mechanism, and validated by independent raters with inter-annotator agreement reported. We pair the dataset with a leak-aware protocol that removes a near-perfect onset leak, and a baseline study across classical, recurrent, and attention-based model families. Under this honest protocol, drift is identifiable and attributable well above random and majority baselines across every family (affected layer macro-F1 near 0.70, responsible actor near 0.85, causal mechanism near 0.49), and heavy attention offers no advantage over simpler models on this symbolic benchmark.**

## I. Introduction

Modern autonomous systems rarely act alone. Human operators, AI agents, and automated controllers increasingly share control in tightly coupled feedback loops, each responding to the others many times within a single task. As these systems grow more capable and complex, they also grow more prone to *drift*: silent deviations from intended behavior that accumulate unnoticed until something goes wrong [7], [8]. Because no alarm sounds and no error is thrown, drift is especially dangerous in the high-stakes settings where these hybrid systems are deployed: clinical decision support, power-grid operations, autonomous driving, and industrial automation[4].

Despite these stakes, the means for studying drift have not kept pace. Existing drift-detection benchmarks focus largely on single-modality, single-actor settings [2], [8]: one sensor stream, one model, one decision-maker. They offer little traction on the layered, multi-actor stacks of real deployments, where a disturbance can arise in any layer and be traced to any actor. As a result, the community lacks a controlled testbed for studying not merely whether drift can be detected, but where it began, who or what caused it, and how to recover.

Building such a testbed is itself difficult, for four reasons. First, the relevant signals are fragmented: human actions, AI predictions, controller commands, sensor readings, active constraints, and failures are typically logged separately and rarely time-aligned. Second, failures are hard to label, because a single breakdown may originate in drift, model uncertainty, controller limits, a delayed human response, or some combination. Third, a benchmark must be realistic yet safe, reproducing high-risk failures without endangering a live system. Fourth, evaluation is hard: success turns not on accuracy alone, but also on resilience, recovery time, the quality of human intervention, constraint satisfaction, and safe degradation under stress.

This work closes that gap by contributing five artifacts to the community, summarized as follows:

- A public, time-aligned multi-layer trace dataset with ground-truth labels for drift type, affected layer, onset time, responsible actor(s), and causal mechanism.
- A leak-aware evaluation protocol that separates genuine learning from memorization of the synthetic injection, so reported scores reflect real capability.
- A repair and audit-ready label schema that supports tasks beyond detection without re-engineering the dataset.
- A cross-family baseline study showing that classical, recurrent, and attention-based models all learn drift well above chance under the honest condition, evidence the benchmark's signal is genuine.
- A cost-efficiency analysis identifying compact recurrent encoders (GRU and LSTM) as the most efficient baselines, matching or exceeding the Transformer at lower cost.

## II. Related Work

### A. Drift

Concept drift refers to an online supervised-learning scenario in which the relation between the input data and the target variable changes over time [8]. The Gama survey characterizes it as a change, between two points in time, in the joint distribution of a system's inputs and labels, so the same input can yield a different outcome as conditions evolve [8]. That survey frames detection around three requirements: a single model with a single output, a continuous stream of ground-truth labels from which an error rate is computed, and ongoing monitoring of that error rate until it climbs high enough to declare drift.

Several methods instantiate this paradigm. The Drift Detection Method monitors a classifier's online error rate and signals drift when it exceeds a statistically derived threshold [7]. Kifer et al. [3] compare a reference window against a recent one with a two-window distribution-distance test. Barddal et al. [2] survey feature drift, where the relevance of individual features changes over time. Adams and MacKay [1] cast change-point detection as Bayesian inference over the time since the last change. Aguiar and Cano [15] analyze how the locality of drift within a stream affects detector behavior.

Collectively, [8], [7], [3], [2], and [1] establish drift detection as a single-stream, single-model paradigm that monitors one model's error rate against its ground-truth labels. This framing leaves three gaps that the present work addresses. First, these methods detect that drift occurred but not where it originated; assuming no layered system in which state, perception, decision, policy, or actuation can each drift independently, they cannot attribute a deviation to a layer, actor, or causal mechanism. Second, they wait for the error rate to accumulate, whereas the per-step ground truth used here allows drift to be flagged the moment it begins. Third, they are single-modality monitors, whereas coordination failures in human-AI-controller systems are rarely confined to one channel. Our per-step model therefore inherits the formal drift definition and change-point primitives of this literature and extends them across two new dimensions: system layer and actor type.

### B. *Available Similar Benchmarks*

The benchmark is derived from ALFRED [11], whose 25k expert demonstrations pair planner-generated optimal trajectories with crowd-worker language directives, cross-validated by multiple Amazon Mechanical Turk annotators, as in other crowd-validated benchmarks [12], giving a clean, high-quality base from which to inject controlled drift. TEACh [9], the closest published neighbor, established the importance of multi-actor, correction-aware embodied benchmarks; this paper extends its two-human Commander/Follower split to three actor types: human, AI agent, and automated controller.

A second, recent line introduces benchmarks adjacent to ours in the agent setting. Zhang et al. [17] release Who & When for automated failure attribution in multi-agent LLM systems, asking which agent and which step caused a failure, and report that even the strongest methods remain far from reliable. TrajAD [16] localizes trajectory anomalies and their onset in LLM agents [13]. These share our interest in locating where and when something goes wrong, but they operate on free-text or single-channel trajectories rather than time-aligned, multi-layer execution traces, and none provide closed-set labels for drift type, affected layer, responsible actor, causal mechanism, and onset on the same data. We share only binary drift detection with this literature; the per-layer, per-actor, and per-mechanism attribution that distinguishes this benchmark has no published peer. Because no prior work reports this task, a head-to-head numerical comparison is not meaningful, so Table 1 positions our work against these adjacent efforts as reference points, each measured on its own benchmark.

| Work | Task / data | Result |
|---|---|---|
| Drift detectors [7], [3], [1] | stream drift detection (SEA, Electricity) | acc ≈ 0.82–0.85 [15] |
| Zhang et al. [17] | failure attribution: agent + step (Who&When) | 53.5% / 14.2% |
| TrajAD [16] | trajectory anomaly + onset (TrajBench) | beats baselines |
| *This work* | *detection + layer / actor / mech / onset* | *layer 0.71, actor 0.84; onset 4.1* |

Table 1. Positioning against adjacent prior work.

## III. Methodology

### A. *Dataset*

The benchmark is created synthetically: controlled drift is injected into ALFRED expert demonstrations of everyday household tasks, and every label is then screened by an automated validator and audited by human raters before entering the dataset (Figure 1). The procedure yields 1,918 drifted traces split 1,342 / 287 / 289 (70/15/15) under split seed 42, one trace per demonstration; nominal traces are also generated and mixed into the splits as drift-free negatives for the drift-presence head.

*Generation.* One of five drift-injection subroutines, one per drift type (world, perception, policy, constraint, human override), is applied to a nominal trace; the types span ten causal mechanisms. Independently, five system layers (state, observation, decision, rules, control) record where the deviation surfaces, and a per-step actor field records who acted (human, AI agent, automation, or policy layer at injection points; the automated controller otherwise). Each drift label records the drift type, drift layer, intervention time, and repair episode.

Injection is relevance aware. The intervention time is drawn uniformly between 20% and 80% of the trace length (never before step 5), and an injection is accepted only if it produces at least one observable post-intervention effect on behavior; otherwise the injector re-rolls the drift type for that trace. This filter is why world drift is slightly under-represented (318 traces versus 400 for the other types): it requires an object the agent later acts on. Concretely, world drift teleports such an object to a randomly chosen receptacle, so later interactions with it fail as unreachable; perception drift either swaps the perceived identity of the acted-on object or inserts a hallucinated object that the agent then targets; policy drift overwrites a ten-step window with an action loop, wrong actions, or excessive rotation; constraint drift activates one of four rules (no knives allowed, bathroom restricted, fridge locked, stove disabled) selected so that it blocks at least one later action; and human override hands control to a scripted human for three to eight steps whose actions visibly diverge from nominal, followed by an explicit handback step and post-override resynchronization flags.

Because each subroutine knows exactly what it changed, every label field (drift type, affected layer, onset, responsible actor, causal mechanism) is recorded deterministically at injection time rather than annotated post hoc, and a rule-based repair simulator attaches a repair episode whose distance-to-nominal is the normalized Hamming distance between the drifted and nominal action sequences. Figure 2 walks one such drifted trace end to end.

| Drift type | Layer | Definition | Examples |
|---|---|---|---|
| World | State | Object positions or states change unexpectedly in the environment | Object teleported, receptacle toggled open/closed, object state mutated |
| Perception | Observation | The agent's observations become noisy or incorrect | Sensor misidentifies an object, object occluded from view |
| Policy | Decision | The agent's action-selection patterns shift from expected behavior | Action loop, wrong action selected, excessive rotation |
| Constraint | Rules | Active policy or security rules change mid-episode | New constraint injected that restricts previously valid actions |
| Human Override | Control | Control authority transfers from the AI agent to a human operator | Human intervenes to correct agent behavior or take over task |

Table 2. Drift Type Definitions and Examples

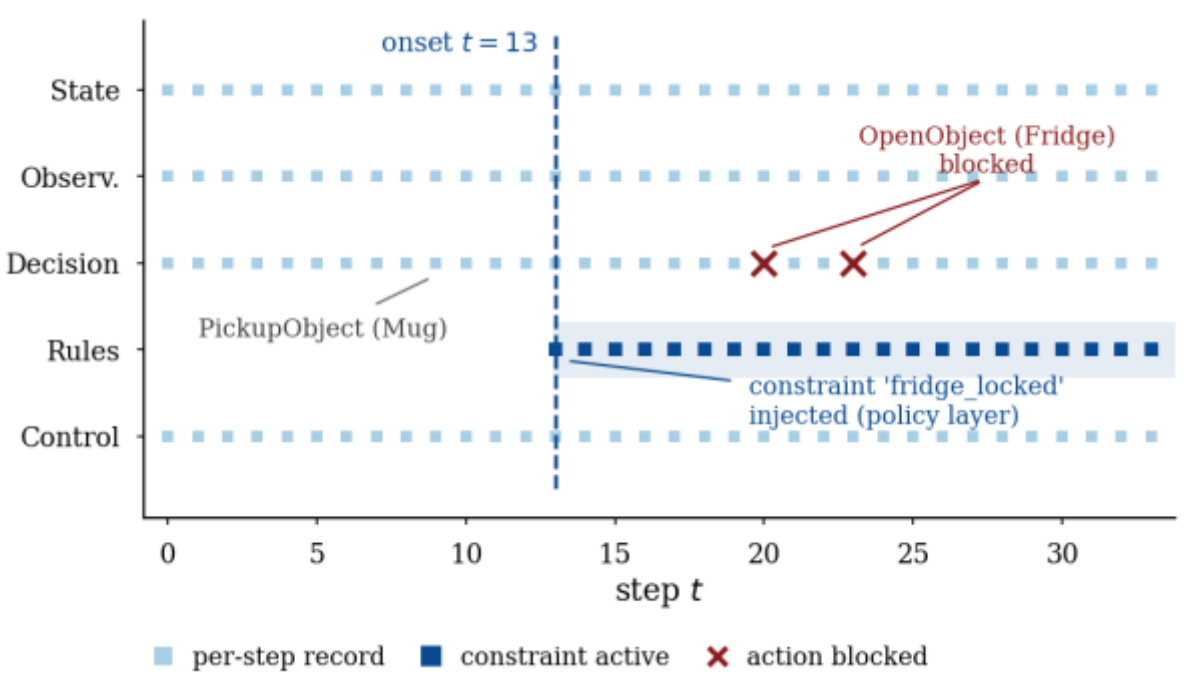

Figure 2. Example drifted trace (constraint type): the fridge_locked rule is injected at step 13 (rules layer) and the agent's fridge interactions at steps 20 and 23 are blocked (decision layer).

*Validation.* Every label passes a two-stage, human-in-the-loop check. An automated four-layer validator screens structure, attribution, repair episode, and behavior, and a record must pass all four to reach human review. Raters then audit each surviving trace against its drift label in a .NET Blazor tool, the Drift Audit Tool, marking it correct, incorrect, or uncertain, with an active learner serving the most uncertain records first.

Agreement is reported in aggregate (Cohen's/Fleiss' κ), and records clearing both stages form the curated dataset.

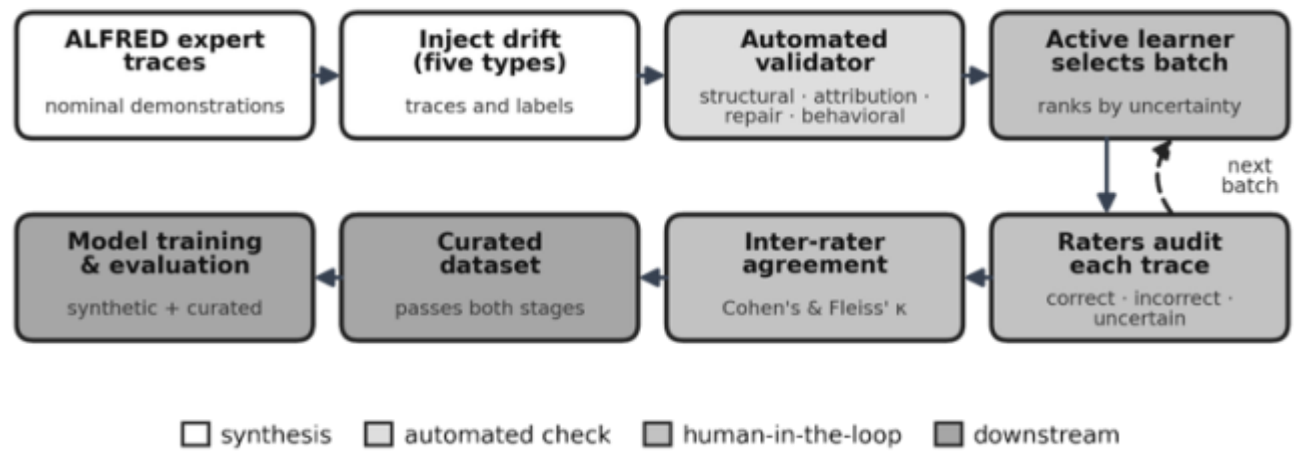

Figure 1. Dataset Generation & Validation Procedure

## B. Model Architecture

Each trace emitted by the preceding pipeline stages is encoded as a sequence of per-step records, each described by eight input features. Each *TraceStep*, the per-step object produced upstream by the drift injector, is converted into a vector by embedding its categorical fields (action, object, actor, subgoal) and encoding its structural fields (constraints, state deltas), then projected to a 128-dimensional representation. These per-step vectors pass through a sequence encoder that contextualizes each step against the rest of the trace. We treat this encoder as a swappable component and instantiate it as a bidirectional GRU, a bidirectional LSTM, or a Transformer, so that all learned models share the same inputs, prediction heads, and training pipeline and differ only in how they encode sequence context. The Transformer instantiation adds a sinusoidal positional encoding and a four-layer, pre-norm encoder [14] with GELU activations and 0.1 dropout, mixing information across steps through self-attention; the recurrent instantiations carry context through bidirectional recurrence.

Four trace-level heads (drift presence, drift layer, primary actor, and causal mechanism) read a mean-pooled summary of the encoder output to predict labels for the whole trace, while a fifth head determines drift onset step-by-step, flagging the first step whose drift-presence probability surpasses 50%.

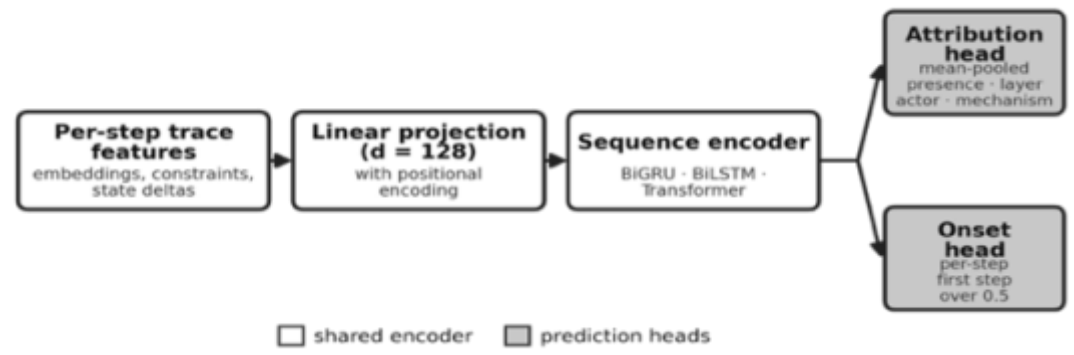

Figure 3. Model Architecture (rendered horizontally)

## C. Training Pipeline

An AdamW optimizer with a learning rate of $3e-4$ and weight decay of 0.01 was set as hyperparameters, with a linear warmup of 500 steps followed by cosine decay, gradient clipping at norm 1.0 and mixed-precision arithmetic on a single NVIDIA A100 GPU. The model was trained for 15 epochs at a batch size of 32 traces using a uniformly weighted sum of cross-entropy losses for the four attribution heads and length-masked binary cross-entropy for the onset head. The data split is fixed by seed 42, and each model is trained over three seeds (42, 43, 44), with reported metrics given as the mean and standard deviation across them.

## D. Baseline

This paper reports its metrics alongside three baselines on the held-out 289-drifted-trace test split: a random baseline that predicts uniformly at random per head; a majority-class baseline that always predicts the mode of the training-set labels; and a rule-based leak baseline that flags the first step with a non-empty realized state delta as the predicted intervention. The rule-based baseline serves a specific purpose, quantifying the dataset's onset shortcut so the leaky run's near-perfect onset numbers can be recognized as imitation of this rule rather than genuine learning.

To probe whether the benchmark is solvable across architectures, the baseline suite spans three model families: classical order-free classifiers over a bag of symbolic trace features (logistic regression, random forest, and gradient-boosted trees [17]), bidirectional recurrent encoders (a BiGRU and a BiLSTM), and the multi-task Transformer described above. All learned models share identical inputs, the same prediction heads, and the same training and evaluation pipeline; each is trained over three random seeds (42, 43, 44) under the honest condition, with the onset decision threshold selected on the validation split. For each model the suite also records cost, namely trainable parameters, training time, and inference throughput, so that accuracy can be weighed against computational expense.

## IV. Experimental Configuration

### A. Dataset Design

The benchmark defines five drift types, each anchored to one system layer and to the actor responsible for it. See Table 2. Drift type maps one to one onto system layer, while the responsible actor is assigned per causal mechanism rather than per type, so perception splits between AI agent (misidentification) and automation (occlusion). The five types span ten mechanisms in all, with constraint drift governed by an explicit rule set checked by the behavioral validator.

Although injected synthetically, each drift type is patterned on a documented real-world failure mode: world drift mirrors unmodeled environment changes such as objects moved by another actor; perception drift mirrors sensor faults, misidentification, policy drift mirrors behavioral regressions after a model update or fine-tuning; constraint drift mirrors rules and safety interlocks imposed mid-operation; and human override mirrors manual takeover in semi-autonomous operation. The benchmark trades observational realism for exhaustive, controlled labels; the Limitations section states the boundaries of that trade.

The dataset in this paper comprises 1,918 total drifted traces with a split size of 1,342 records for training, 287 records for validation, and 289 records for testing (70/15/15). A seed split of 42 was used to keep sequential runs deterministic while the dataset was derived from the ALFRED expert demonstrations. A single ALFRED demonstration equals one drifted trace. Nominal traces are also produced by the trace generator, and are mixed into the train, validation, and test splits as drift-free negatives for the drift-presence head (the test split contains 289 drifted plus 150 nominal traces); the attribution and onset metrics reported here are computed on the drifted traces only. See Table 3.

| **Drift type** | **Traces** | **%** |
|---|---|---|
| World | 318 | 16.6 |
| Perception | 400 | 20.9 |
| Policy | 400 | 20.9 |
| Constraint | 400 | 20.9 |
| Human override | 400 | 20.9 |
| **Total** | **1,918** | **100** |

Table 3. Distribution of Traces per Drift Type

The dataset also tracks who, the responsible actor, was the cause of the drift. The paper defines four responsible actors: human, AI agent, automation, and policy layer. See Table 4.

| **Responsible actor** | **Traces** | **%** |
|---|---|---|
| AI agent | 667 | 34.8 |
| automation | 451 | 23.5 |
| human | 400 | 20.9 |
| policy layer | 400 | 20.9 |
| **Total** | **1,918** | **100** |

Table 4. Percentage of Traces Per Responsible Actor

### B. Evaluation Metrics

Evaluation uses a two-condition design on the same held-out test split of 289 drifted traces: a 'leaky run' where all features are present, and an 'honest run' where the realized state delta field is zeroed out to determine whether the model is genuinely learning to detect drift. Without the honest run, the model would have simply learned to read the realized state delta flag, which the drift injector populates at and after the intervention step by construction, instead of learning when drift began, which system layer was affected, and which actor was responsible. Essentially, the model would simply learn to read the realized state delta flag instead of learning to infer when, what system, and who was responsible for said drift if this field were not zeroed. The honest condition removes only this single field; all other inputs (actions, objects, actors, subgoals, and constraint context) remain available, so models keep the full behavioral signal while losing the injector's bookkeeping. Macro-F1 and accuracy are the reported metrics for the three trace-level attribution heads (drift presence, drift layer, primary actor, causal mechanism); for the per-step onset head the paper reports MAE and exact-hit rate against the ground-truth intervention time field in the *drift label,* the integer step index at which the drift injector intervened in the trace.

### C. Implementation

All experimentation was implemented and conducted in Google Colab using PyTorch 2.x on a single NVIDIA A100 GPU. Drift generation is contained to a single Jupyter notebook, comprised of five drift-injection methods, one per drift type. The training pipeline loads the dataset from a Google drive folder via a service-account authentication and completes a 15-epoch run. A separate .NET Blazor app/utility handles manual annotation validation.

The public release comprises one JSONL trace file and one JSON drift-label file per drifted trace, together with fixed train/validation/test manifests for the 70/15/15 split (seed 42) so that results can be reproduced without re-splitting. The generation, training, and evaluation notebooks and the Drift Audit Tool are released under the MIT license, with the dataset under CC BY 4.0, and usage notes document trace loading and the honest-condition feature masking. All artifacts are available at github.com/m0nkspade/TRACE-benchmark.

## V. Experimental Results

### A. Annotation Synthesis

The procedure used in this paper synthetically generated 1,918 drifted traces along with their respective drift labels (*drift labels)*. The process comprises a method that randomly selects from five subroutines, each intended to inject one of the five drift types defined in this paper. The five drift types collectively span ten causal mechanisms: object teleported, object state changed, receptacle toggled (within the world drift type); sensor misidentification, object occlusion (within the perception drift type); action loop, wrong action, excessive rotation (within the policy drift type); added constraint (within the constraint drift type); and human intervention (within the human override drift type). Each *drift label* consists of fields for drift type, drift layer, intervention time, repair episode; the paired trace additionally records a per-step actor field (*automated controller at the agent's own steps; one of human, AI agent, automation, or policy layer at drift-injection points*) indicating who took each step.

### B. Annotation Validation

Validation of drift labels is a two-stage process. The automated stage applies a four-layer validator, and a record must pass all four to enter the dataset: a structural check that required fields are present and drawn from the controlled vocabulary, an attribution check that the responsible actor matches the injected drift type, a repair-episode check that each repair step is filled out and well-formed, and a behavioral check that the trace evidence is consistent with the declared drift type (for constraint drift, that a constraint-violating step actually appears in the trace).

The second stage is a .NET Blazor tool for manually inspecting each trace against its drift label. Raters audit an independently sampled, private batch assigned only to them; individual ratings stay private and are reported only in aggregate, as an inter-rater agreement report (Cohen's κ for pairs, Fleiss' κ for three or more raters) used to check annotation reliability.

### C. Inter-Annotator Agreement

We assess label reliability through inter-annotator agreement: three raters independently audited held-out traces in the Drift Audit Tool, confirming or rejecting each proposed label without seeing the others' verdicts. We report raw percent agreement, Cohen's κ [5] per rater pair, and Fleiss' κ [6] across all three.

Across 470 verdicts, raters confirmed the label accurate in 98.1% of traces and flagged none as inaccurate, with raw agreement of 90 to 100% on co-reviewed traces (100% on a 75-trace overlap, 90% on a 50-trace overlap). Per-session Fleiss' κ ranged from 0.45 (moderate) to 1.0.

The chance-corrected coefficients need care: with 98.1% of verdicts in one category, expected agreement is near unity, so Cohen's and Fleiss' κ are deflated by the high-prevalence paradox, where κ can approach zero despite over 90% raw agreement. We therefore read raw agreement and the verdict distribution as the primary evidence and κ as a conservative lower bound, indicating the injected drift signals are unambiguous to human readers and the training and evaluation labels are reliable.

The audit evidence spans the full label space. Table 9 breaks the 470 verdicts down by drift type: raters audited 203 distinct drifted traces (10.6% of the corpus, with 9.0% to 12.6% coverage of every type) and confirmed the injected label in 91.8% to 100% of verdicts per type, with no label judged incorrect. All nine uncertain verdicts concentrate in world drift (six) and constraint drift (three). Raters additionally audited 46 nominal traces as negative controls and unanimously confirmed all 93 of those verdicts as drift-free, so the audit checked both that injected drift is visible and that its absence is not over-reported. Among the 137 drifted traces reviewed by two or more raters, verdicts were unanimous on 100% of perception, policy, and human-override traces, 93.3% of constraint traces, and 87.5% of world traces, mirroring the model-side finding that state-layer (world) drift carries the subtlest signal (Figure 4).

| Drift type | Verdicts | Traces | Coverage | Confirmed | Uncertain |
|---|---|---|---|---|---|
| World | 73 | 40 | 12.6% | 91.8% | 8.2% |
| Perception | 64 | 36 | 9.0% | 100% | 0% |
| Policy | 67 | 40 | 10.0% | 100% | 0% |
| Constraint | 78 | 39 | 9.8% | 96.2% | 3.8% |
| Human override | 95 | 48 | 12.0% | 100% | 0% |
| Nominal (control) | 93 | 46 | n/a | 100% | 0% |

Table 9. Human audit verdicts by drift type (three raters); nominal traces are drift-free negative controls.

### D. Results

As affirmed by Table 5, the model was in fact exploiting the leak caused by training on the full set of features: drift layer drops by 11.4% and causal mechanism by 23.6%. The leaky run is not a true measure of performance, since training on the full feature set is akin to handing the model an answer key; with the leak removed, the honest run still performs well, and the gap between the two confirms that the leaky feature was substantially and artificially inflating the model's performance.

| Head | Metric | Leaky | Honest | Δ |
|---|---|---|---|---|
| **drift layer** | macro-F1 | 0.82 | 0.706 | **−0.114** |
| **primary actor** | macro-F1 | 0.85 | 0.848 | **−0.002** |
| **causal mechanism** | macro-F1 | 0.65 | 0.414 | −0.236 |
| **onset** | MAE (steps) | 0.06 | 8.39 | **8.33** |
| **onset** | exact-hit rate | 0.98 | 0.218 | **−0.762** |

Table 5. Leaky Run vs. Honest Run

What is noteworthy is that every model we trained clears the sanity baselines: across the classical, recurrent, and attention-based families, all of them beat the random and majority-class baselines on every attribution head by a sizable margin. These margins reflect genuine, learnable drift structure rather than the onset leak. See Table 6 and Table 7.

| Head | Random | Majority | Honest model |
|---|---|---|---|
| drift layer macro-F1 (5-way) | *≈0.20* | *≈0.25* | 0.706 |
| primary actor macro-F1 (4-way) | *≈0.25* | *≈0.40* | 0.848 |
| causal mechanism macro-F1 (10-way) | *≈0.10* | *≈0.15* | 0.414 |
| onset MAE (steps) | *≈ T/3* | *≈ mean onset* | 8.39 |
| onset exact-hit rate | *≈ 1/T* | *small* | 0.218 |

Table 6. Honest Run vs. Sanity Baselines

| Method | layer F1 | actor F1 | mech F1 | composite |
|---|---|---|---|---|
| Logistic regression | 0.702 | 0.86 | 0.488 | 0.784 |
| Random forest | 0.693 | 0.814 | 0.494 | 0.768 |
| Hist. gradient boosting | 0.700 | 0.840 | 0.510 | 0.781 |
| BiGRU | 0.736 ± 0.031 | 0.866 ± 0.017 | 0.461 ± 0.097 | 0.803 ± 0.027 |
| BiLSTM | 0.758 ± 0.011 | 0.867 ± 0.006 | 0.505 ± 0.004 | 0.811 ± 0.002 |
| Transformer (multi-task) | 0.710 ± 0.009 | 0.849 ± 0.007 | 0.483 ± 0.018 | 0.790 ± 0.004 |

Table 7. Cross-family comparison on the honest test split (mean ± SD over three seeds; classical rows shown as means).

We evaluate the three baseline families on the held-out test split under the honest condition, reporting per-head macro-F1 and onset error as the mean and standard deviation over random seeds.

**Dataset utility across model families.** Architectures as different as gradient-boosted trees, recurrent networks, and self-attention all recover the drift labels well above chance on every head, but only once the synthetic onset leak is withheld. This points to genuine, model-agnostic structure rather than an artifact any single architecture exploits, with mechanism attribution the hardest head.

**Cost efficiency.** On a fair budget the compact bidirectional recurrent encoders are the most accurate models. The BiLSTM tops the attribution composite (0.811 ± 0.002) and the BiGRU follows within seed noise (0.803 ± 0.027) while remaining the smallest neural model (0.63M parameters versus 0.81M for the Transformer) and posting the lowest onset MAE (4.1 ± 0.6 steps). The Transformer sits at the classical floor (composite 0.790 ± 0.004 against 0.784 for logistic regression), so attention buys no advantage on this symbolic benchmark. The classical baselines are the cheapest to train, remain within a few points on attribution, are deterministic across seeds, and still take the exact-onset head, making them a strong low-cost reference.

We therefore report the compact bidirectional recurrent encoders as the headline baselines for this benchmark: the BiLSTM is the most accurate and most stable on the attribution composite, and the BiGRU matches it within seed noise while using the fewest parameters of any neural model and achieving the lowest onset error. Recurrence helps where attention does not, and the Transformer falls to the classical floor.

Figure 4 shows the pooled test confusion matrices of the bidirectional GRU. Errors concentrate almost entirely in two adjacent pairs: state versus observation for the affected layer, and AI agent versus automation for the responsible actor; the rules, control, human, and policy-layer classes are recovered nearly perfectly. Attribution difficulty is therefore localized to the two structurally closest class pairs rather than spread across the label space.

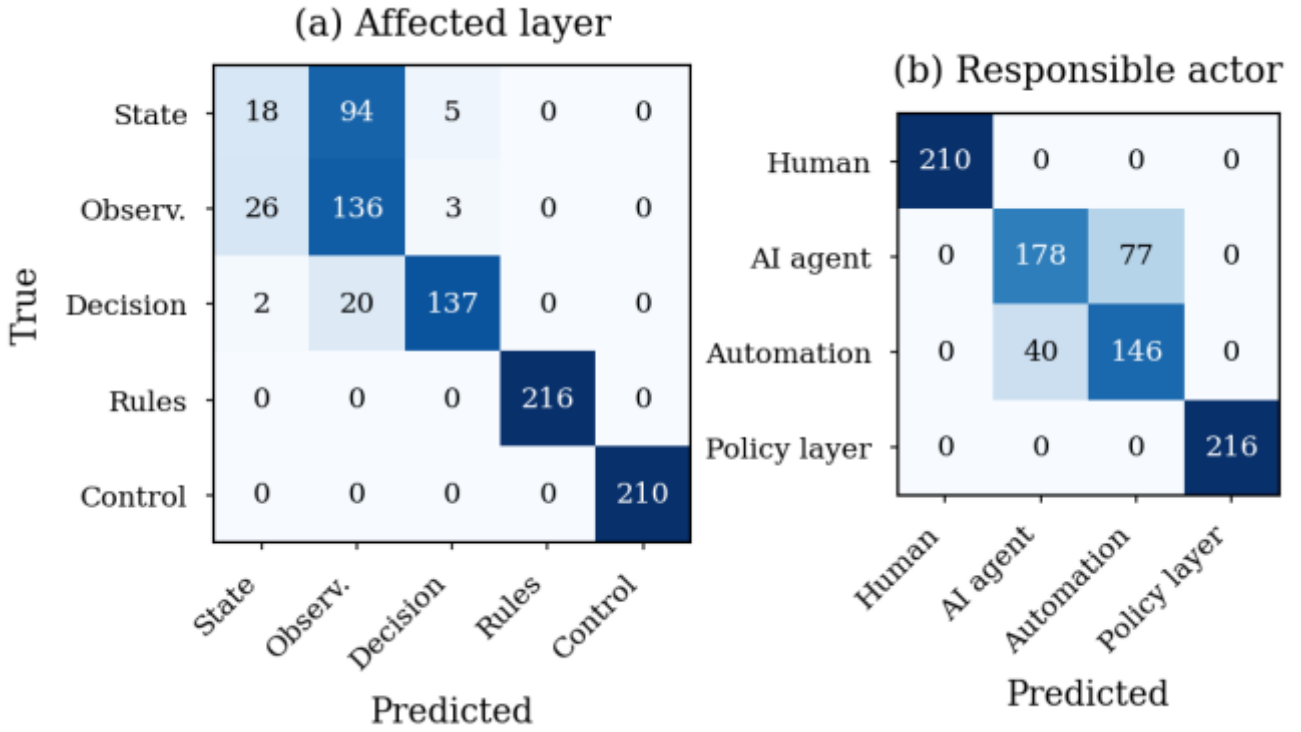

Figure 4. BiGRU confusion matrices, honest protocol, pooled over three seeds: (a) affected layer, (b) responsible actor; shading row-normalized, cells show counts.

**Onset tolerance windows.** Exact-step onset matching is a strict criterion, so Table 8 also reports hit rates within ±1, ±3, and ±5 steps of the true intervention step. The bidirectional GRU localizes onset within one step in 52% of drifted test traces and within three steps in 67%, indicating that the remaining onset error is dominated by near misses rather than gross mislocalization.

| Model | MAE (steps) | hit@0 | hit@±1 | hit@±3 | hit@±5 |
|---|---|---|---|---|---|
| BiGRU | 4.07 ± 0.55 | 0.354 ± 0.060 | 0.521 ± 0.103 | 0.667 ± 0.071 | 0.750 ± 0.044 |
| BiLSTM | 4.41 ± 0.21 | 0.311 ± 0.060 | 0.581 ± 0.040 | 0.696 ± 0.003 | 0.753 ± 0.007 |
| Transformer | 5.03 ± 0.21 | 0.428 ± 0.010 | 0.494 ± 0.019 | 0.602 ± 0.025 | 0.677 ± 0.025 |

Table 8. Onset localization: hit rate within ±k steps of the true onset (mean ± SD over three seeds).

### E. *Active Learning: Label Efficiency*

Section III described the active-learning queue that orders traces for human review by model uncertainty; here we test whether that ordering reduces annotation effort. Using a random forest over per-step symbolic features as a fast surrogate, we grow the labeled set in fixed increments, choosing each batch either by predictive uncertainty (entropy) or at random from a pool of 1,342 traces. Attribution quality is the mean macro-F1 across the layer, actor, and mechanism heads on the 289 held-out test traces, averaged over three seeds (Figure 5).

Attribution quality climbs steeply over the first few hundred labeled traces and plateaus near 0.66 macro-F1, recovering most of the full-budget quality from under a quarter of the traces. The strategies separate where it matters most, in the low-label regime: uncertainty sampling reaches 0.65 macro-F1 after 300 labeled traces, while random selection needs 400. Active learning thus attains equal quality from 75% of the labels, a 25% reduction in verification effort, before the curves converge. The effect is modest but confirms that prioritizing uncertain records returns more labeling value per trace.

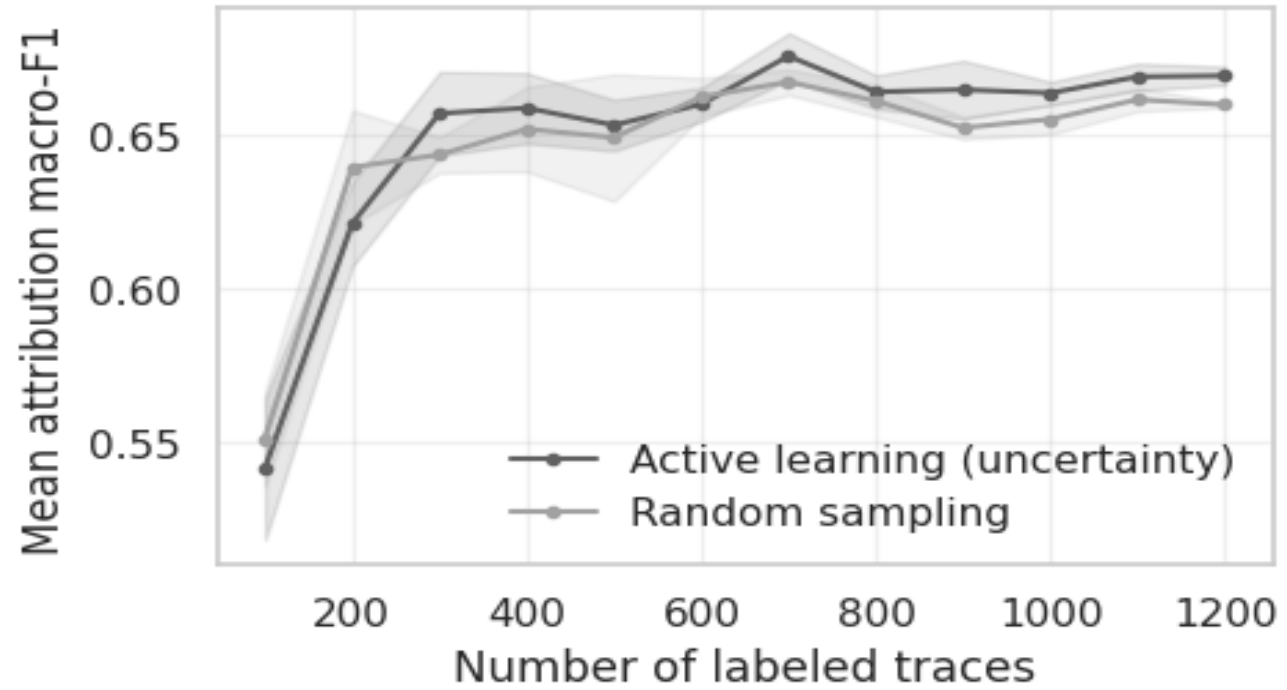

Figure 5. Label efficiency: mean attribution macro-F1 versus labeled-set size for uncertainty-based and random selection (mean ± SD, three seeds).

## VI. Limitations

The design choices that make the benchmark controllable also bound its claims, in four ways. First, all drift is synthetically injected and every attribution label is generated by the injector rather than observed in a deployed system: label fidelity is exact by construction, but each injected signature is an idealized instance of its mechanism, cleaner than the mixed, gradual deviations of practice. Second, the corpus derives from a single domain (ALFRED household tasks) and totals 1,918 traces, so transfer to industrial control, clinical, or driving settings is untested, and claims should be read as scoped to symbolic, episodic traces of this kind. Third, human validation established that the injected labels are plausible and recoverable from trace evidence; it did not test whether benchmark performance predicts real operator diagnosis, which requires a human-in-the-loop study. Fourth, because drift type maps one to one onto the affected layer, type and layer prediction are not independent tasks, and mechanism attribution (the hardest head) is the more informative measure of fine-grained understanding. We therefore position this benchmark as a controlled testbed for developing attribution methods before they meet observational data, not as a substitute for real-world evaluation.

## VII. CONCLUSION

We introduced a multi-layer trace benchmark for drift detection and attribution in systems where humans, AI agents, and automated controllers share control, built by injecting controlled drift into ALFRED demonstrations and labeling each step across five execution layers with its drift type, affected layer, onset, responsible actor, and causal mechanism. Across classical, recurrent, and attention-based families, models identify the affected layer, responsible actor, and causal mechanism well above random and majority baselines (macro-F1 near 71%, 85%, and 41 to 51% respectively), localizing drift onset to within four to nine steps depending on the model family. No single architecture dominates: a linear classifier matches the neural models, so heavy attention buys no accuracy on this symbolic benchmark, and fine-grained mechanism classification remains the open frontier. Our central methodological contribution is a leak-aware protocol that zeroes the realized state delta field, a near-perfect onset leak the injector creates by construction, and compares this honest run against a full-feature upper bound, yielding a principled measure of how much reported performance is real rather than an artifact of synthetic injection, one that generalizes to any similarly constructed benchmark.

## VIII. FUTURE WORK

Two directions extend the benchmark from inferring drift to acting on it. Its repair episode labels (corrective action, phase progression, distance to nominal, and outcome) already support training repair models that predict corrective actions and their feasibility, with no new data collection. Building on this, a closed-loop dispatcher could use detection, attribution, and repair-feasibility signals to decide when to act, which agent or human to invoke, and how to verify the correction, under safety and audit constraints.